\documentclass[lettersize,journal]{IEEEtran}
\usepackage{amsmath,amsfonts}
\usepackage{algorithmic}
\usepackage{algorithm}
\usepackage{array}
\usepackage[caption=false,font=normalsize,labelfont=sf,textfont=sf]{subfig}
\usepackage{textcomp}
\usepackage{stfloats}
\usepackage{url}
\usepackage{verbatim}
\usepackage{graphicx}
\usepackage[table]{xcolor}
\usepackage{xcolor}
\usepackage{cite}
\usepackage{amsmath}
\usepackage{multicol}
\usepackage[square, comma, sort&compress, numbers]{natbib}
\usepackage{multirow}
\usepackage{booktabs}
\usepackage{pifont}
\usepackage{xcolor}
\usepackage{caption}
\usepackage{subcaption} 
\usepackage[table]{xcolor}
\usepackage{booktabs}

\usepackage{siunitx}
\usepackage[colorlinks=true,       
            linkcolor=blue,        
            citecolor=blue,        
            urlcolor=cyan,         
            bookmarks=true,        
            pdfstartview=FitH]{hyperref}

\begin{document}
\title{InterOCF: Spatio-Temporal 2D-3D Interaction for Camera-Only 4D Occupancy Forecasting}

\author{
\IEEEauthorblockN{Qi Zhang$^{1}$, Xinquan Yu$^{1}$, Kaiyi Zhang$^{1}$, Hui Huang*$^{1}$\IEEEcompsocitemizethanks{
    \IEEEcompsocthanksitem $^{1}$College of Computer Science and Software Engineering, Shenzhen University, Shenzhen, China. Email: qi.zhang.opt@gmail.com, 2410103016@szu.edu.cn, zhangky1999@gmail.com, hhzhiyan@gmail.com
    \IEEEcompsocthanksitem *Corresponding author.
}}}


\maketitle

\begin{abstract}
Camera-only 4D occupancy forecasting enables autonomous vehicles to predict future 3D semantic scenes solely from historical multi-view images, which is critical for driving safety. Even though current methods have achieved good performance, the strong spatial-temporal modeling between the input multi-view frames is still underexplored, which limits the performance of those methods in future 4D forecasting. To address this gap, we introduce a novel framework, InterOCF, for 4D occupancy forecasting that jointly models temporal dynamics in both 3D voxel-based representations and multi-view segmentation sequences, while explicitly incorporating feature interaction between the 2D and 3D branches. Our framework incorporates three core components: 1) A 3D Spatio-Temporal (3DST) module that learns volumetric dynamics from historical voxel states to predict future voxel states; 2) A 2D Spatio-Temporal (2DST) module employing an auxiliary multi-view temporal segmentation forecasting task to enhance temporal semantic dynamics; 3) A Spatio-Temporal Interaction Modeling (STIM) module that enables feature interaction between 2D and 3D representations. Experiments on the nuScenes, Lyft-Level5, and nuScenes-Occupancy datasets show that InterOCF consistently outperforms existing baseline approaches.

\end{abstract}



\section{Introduction}
3D occupancy prediction estimates the 3D semantics of each voxel in the entire scene from input multi-view images, which plays a crucial role in autonomous driving \citep{tong2023scene,ma2024cotr,zhang2023occformer,tian2023occ3d,huang2023tri}. For a better understanding of the dynamic world and providing clues of the future 3D scenes, Camera-only 4D occupancy forecasting (OCF) is further introduced \citep{ma2024cam4docc,chen2025occprophet,9710288,10.1007/978-3-031-19839-7_26,ijcai2023p120,xu2025cvpr,yang2025driving}.

\begin{figure}[t]
\begin{center}
   \includegraphics[width=1\linewidth]{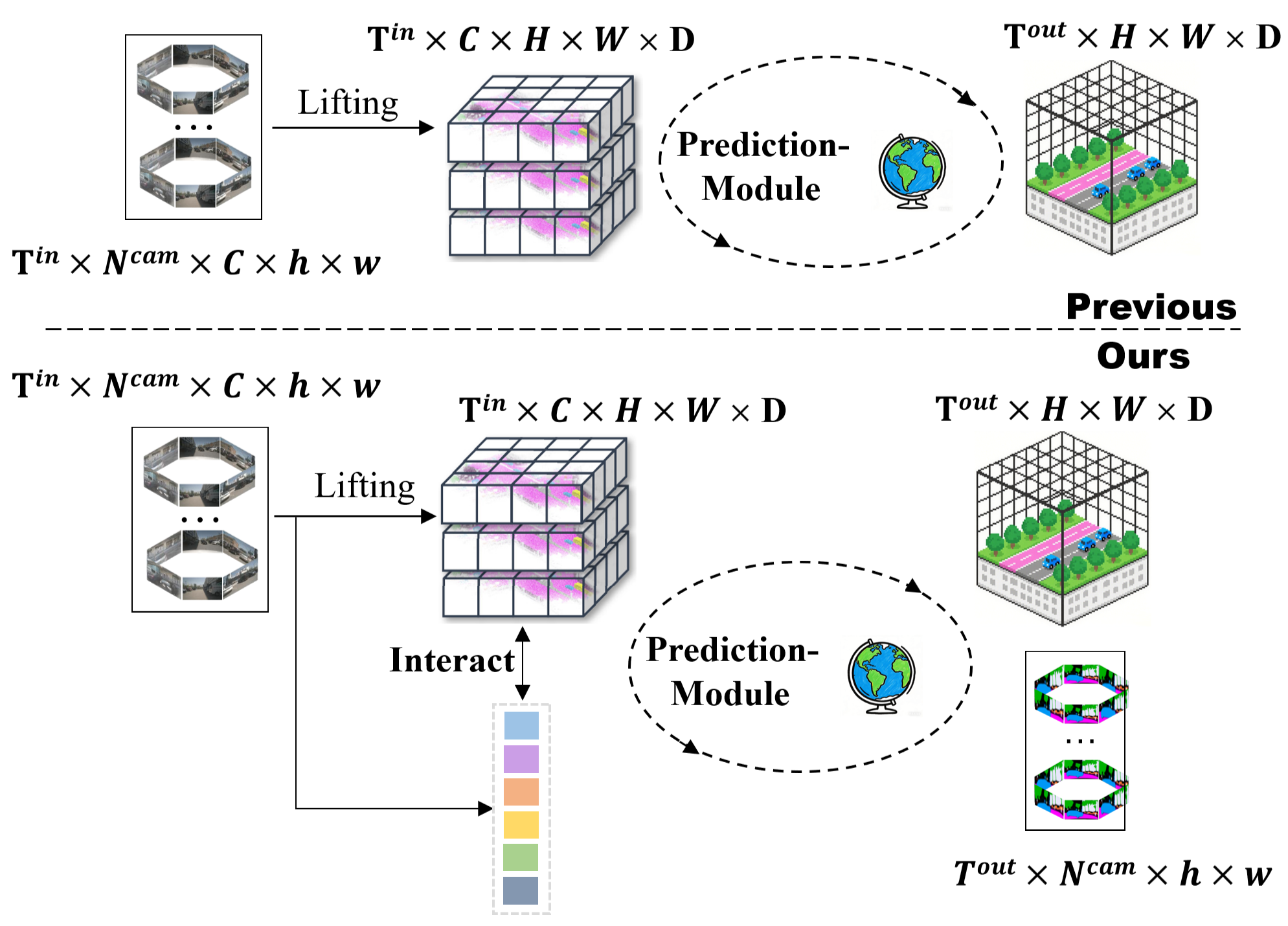}
\end{center}
   \caption{\textbf{Overview of previous vision-based approaches for 4D occupancy forcasting.} Previous vision-based methods for 4D occupancy forecasting typically follow a pure 3D paradigm: they directly map current and past multi-view images to future occupancy predictions. In contrast, our method establishes a 2D-3D interactive paradigm. We simultaneously perform spatio-temporal modeling in both the 2D multi-view image segmentation and 3D occupancy representation spaces, and crucially, we explicitly model and leverage the interaction between them.}
   
 \label{fig:general_idea}
\end{figure}

\begin{figure*}[t]
\begin{center}
   \includegraphics[width=\linewidth]{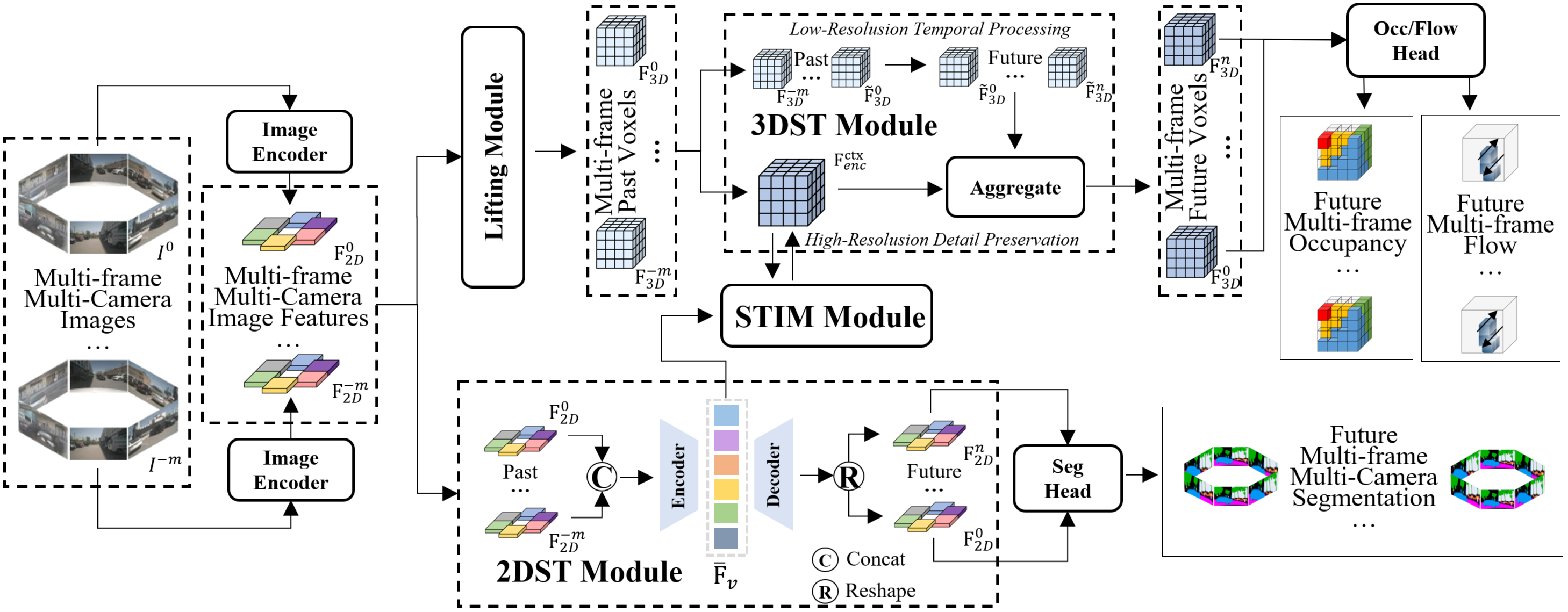}
\end{center}
   \caption{The overall pipeline of our method: First, we extract multi-camera image features from the past and current frames. The bottom 2D branch processes image features through the 2DST module and utilizes a segmentation head to predict future semantic segmentation maps. The upper 3D branch projects image features into 3D space to obtain past voxel sequences. Meanwhile, the fused features from the 2DST and the context features from the 3DST interact through the STIM Module to enhance feature representation. 
   }
 \label{fig:pipeline}
\end{figure*}

Despite the strong performance achieved by recent methods \citep{ma2024cam4docc, chen2025occprophet, xu2025cvpr, yang2025driving}, most of them for 4D occupancy forecasting predominantly adhere to a pure 3D paradigm, as shown in the top part of Fig.~\ref{fig:general_idea}. This paradigm typically takes current and past multi-view images as input to directly predict future occupancy states and the corresponding backward flow. However, this paradigm suffers from several inherent limitations:
(1) Multi-view images are lifted and fused into 3D space only once, after which the image-domain representations are discarded, leading to the loss of rich view-specific cues such as high-resolution appearance details;
(2) conducting temporal roll-out directly in the sparse 3D occupancy space is inefficient and unstable, making long-term forecasting particularly vulnerable to error accumulation;
(3) existing methods fail to explicitly model the mutual interaction between 2D multi-view representations and 3D occupancy dynamics, which is crucial for robust spatio-temporal reasoning in complex driving scenarios.

To address these limitations, we propose a 2D–3D interactive forecasting method, termed InterOCF, which jointly performs spatio-temporal modeling in both the 2D multi-view image space and the 3D occupancy space, as illustrated in the bottom part of Fig.~\ref{fig:general_idea}, and explicitly enables information exchange via dedicated interaction mechanisms between the two branches. 

The complete architecture of our InterOCF is illustrated in Fig.~\ref{fig:pipeline}.
Specifically, we propose a 3D Spatio-Temporal (3DST) module to capture spatio-temporal dependencies in volumetric representations. The module performs temporal modeling on low-resolution volumetric features, where enlarged receptive fields enable more effective capture of global motion patterns in a computation-friendly manner, and subsequently incorporates high-resolution contextual features to preserve fine-grained geometric details.
Furthermore, we introduce a 2D Spatio-Temporal (2DST) module with an auxiliary task that predicts future multi-view image segmentation sequences from historical image features. The segmentation supervision provides rich semantic cues and temporal regularization, complementing the geometric reasoning in 3D space. The semantic segmentation labels are automatically annotated using Segment Anything (SAM) \citep{kirillov2023segment}.
Finally, we introduce the Spatio-Temporal Interaction Modeling (STIM) module to bridge the 2D and 3D representations. The STIM module facilitates bidirectional feature exchange between multi-scale 3D voxels and the 2D branch, enabling cross-modal interaction across these representations. This design enables explicit interaction between the 2D and 3D branches, allowing the model to leverage complementary information from both image space and 3D volume representations. Such cross-modal fusion improves both spatial structure awareness and temporal dynamics modeling, leading to more robust and accurate 4D occupancy predictions.

The contributions of the paper are summarized as follows:

\begin{itemize}
    \item We propose a 3D spatio-temporal (3DST) module that captures global motion patterns via temporal modeling on low-resolution volumes, while incorporating high-resolution features to preserve fine-grained spatial details.

    \item We introduce a 2D spatio-temporal (2DST) module with an auxiliary future multi-view segmentation task, providing additional semantic and temporal supervision to improve scene understanding.

    \item We design a spatio-temporal interaction modeling (STIM) module to bridge 2D and 3D representations, enabling bidirectional feature exchange between multi-view image features and multi-scale voxel representations.

    \item Our InterOCF achieves superior performance across multiple benchmarks, including nuScenes, Lyft-Level5, and nuScenes-Occupancy.
\end{itemize}

\section{Related Work}

\textbf{Camera-Only 3D Occupancy Prediction.}
Camera-Only 3D occupancy prediction \cite{tong2023scene, ma2024cotr, tang2024sparseocc, zhao2024lowrankocc, ouyang2024octocc, zhang2025clip, zhu2024nucraft, lu2024octreeocc, cao2026occany, yan2025pgocc} has attracted considerable attention in autonomous driving for its capability to generate dense 3D environmental representations that surpass traditional bounding box-based methods. Recent advances further extend this paradigm towards more generalized and open-world settings. For instance, OccAny \cite{cao2026occany} explores generalized unconstrained urban 3D occupancy, aiming to improve model robustness across diverse and complex real-world scenarios. Meanwhile, Progressive Gaussian Transformer \cite{yan2025pgocc} introduces an anisotropy-aware Gaussian representation for open-vocabulary occupancy prediction, enabling more flexible semantic understanding beyond predefined categories. To alleviate the heavy reliance on dense 3D annotations, several methods have explored leveraging 2D information as alternative supervision signals. Specifically, Vampire \citep{xu2024regulating} regulates intermediate 3D volume features by incorporating rendered camera-view depth and semantic cues during training. Extending this direction, RenderOcc \citep{10611537} and SelfOcc \citep{huang2024selfocc} adopt more radical strategies that rely exclusively on 2D supervision, completely eliminating the need for 3D labels to reduce annotation costs and complexity. However, these approaches primarily exploit 2D information from a supervisory perspective, which limits its full potential. \textit{In contrast, our approach emphasizes joint 2D and 3D spatio-temporal modeling with explicit cross-branch interaction. This design enables a comprehensive exchange of complementary information, leading to more robust and temporally consistent occupancy predictions.}

\textbf{Camera-Only 4D
Occupancy Forecasting}
Cam4DOcc \citep{ma2024cam4docc} introduces the first benchmark to standardize the evaluation protocol for Camera-Only 4D occupancy forecasting, and proposes OCFNet, an end-to-end network for dense 4D occupancy state forecasting. Drive-OccWorld \citep{yang2025driving} proposed an efficient semantic and motion conditional normalization to enhance the historical BEV feature and conducted temporal cross-attention on it for better temporal modeling. OccProphet \citep{chen2025occprophet} proposed a novel tripling operation to decompose the voxel features into scene, height, and BEV components separately, enabling lightweight spatio-temporal feature interaction and significantly reducing the computational cost. EfficientOCF \citep{xu2025cvpr} decouple spatiotemporal
representation, where BEV occupancy with height values is
utilized for spatial decoupling and instance-aware refinement
in consecutive frames is designed for temporal decoupling. Existing approaches perform 4D spatio-temporal modeling exclusively on the occupancy branch, which limits their ability to exploit rich view-specific and appearance-aware cues and results in insufficient temporal consistency across modalities. \textit{To address this gap, we introduce a novel framework that jointly models temporal dynamics in both voxel-based representations and multi-view image segmentation sequences, enabling complementary spatio-temporal feature learning and more effective cross-domain information exchange.}

\textbf{Spatio-Temporal Modeling in Video Prediction}
Spatio-temporal modeling has been extensively studied in 2D video prediction tasks \cite{guen2020disentangling, wang2020probabilistic, yu2022modular, girdhar2021anticipative, liu2023meta, benaim2020speednet}. Existing approaches explore this problem from various perspectives, including architectural design and semantic modeling. For example, MAU \citep{chang2021mau} introduces an efficient attention-based fusion module to aggregate temporal information across multiple levels, thereby enlarging the temporal receptive field. MotionRNN \citep{wu2021motionrnn} proposes a MotionGRU unit that decomposes object motion into transient variations and long-term motion trends for more accurate prediction. SADM \citep{bei2021learning} further observes that different semantic regions exhibit distinct dynamic patterns and proposes a semantic-aware dynamic model to ensure both geometric and semantic consistency.
More recent works focus on incorporating inductive biases and improving temporal representation learning. PastNet \citep{wu2024pastnet} highlights the importance of physical priors and introduces spectral convolution in the Fourier domain along with a discrete spatio-temporal (DST) module to better capture temporal dynamics. DFDNet \citep{gan2025dfdnet} filters out high-frequency transient noise before modeling temporal dependencies, leading to more stable predictions.
Meanwhile, recent advances extend spatio-temporal modeling beyond conventional video prediction towards more complex reasoning and control-oriented paradigms. Progressive Online Video Understanding with Evidence-Aligned Timing and Transparent Decisions \citep{zhangprogressive} focuses on online video understanding, emphasizing causally consistent reasoning and precise response timing aligned with evidence accumulation in streaming videos. ViPRA \citep{routray2025vipra} further bridges video prediction and control by learning motion-centric latent actions alongside future visual observations, enabling continuous action generation without explicit action annotations.
\textit{In contrast to prior works that primarily focus on 2D video modeling, our approach investigates spatio-temporal modeling in multi-view camera streams and explicitly explores their interaction with corresponding 3D volumetric feature sequences. This joint modeling paradigm enables richer temporal reasoning across both image and 3D space, which is crucial for accurate and consistent 4D occupancy prediction.}

\section{Method}
\subsection{Model Architecture}

As illustrated in Fig.~\ref{fig:pipeline}, our proposed model takes past and current multi-view images $\{I^t\}_{t=-m}^{0}$ as input, where the superscript $t$ denotes the timestamp (with $t=0$ corresponding to the current frame), and $m>0$ represents the length of the historical sequence. First, an image backbone (e.g., ResNet~\citep{he2016deep}) extracts image features $\{F^t_{2D}\}_{t=-m}^{0}$ for each timestamp. These features are then fed into both the 2D branch and the 3D branch, respectively.
In the 2D branch, the image features $\{\mathbf{F}^{t}_{2D}\}_{t=-m}^{0}$ are fed into a \textbf{2D Spatio-Temporal (2DST)} module to predict image features for the current and future time steps, denoted as $\{\mathbf{F}^{t}_{2D}\}_{t=0}^{n}$ ($n>0$ represents the length of the future sequence). The resulting features are subsequently processed by a segmentation head to produce multi-view semantic segmentation maps.
In the 3D branch, we adopt LSS~\citep{philion2020lift} to lift the 2D image features $\{\mathbf{F}^{t}_{2D}\}_{t=-m}^{0}$ into the 3D space, resulting in a sequence of voxel features $\{\mathbf{F}^{t}_{3D}\}_{t=-m}^{0}$. These voxel features are subsequently fed into a novel \textbf{3D Spatio-Temporal (3DST)} module, which models temporal dynamics in the volumetric domain and produces voxel features $\{\mathbf{F}^{t}_{3D}\}_{t=0}^{n}$ for both the current and future time steps. 
Meanwhile, our proposed \textbf{Spatio-Temporal Interaction Modeling (STIM)} 
facilitates cross-domain interaction by integrating the fused features 
$\{\bar{\mathbf{F}}_{v}\}_{v=1}^{V}$ from the 2DST module, where $V$ denotes 
the number of views, with the contextual features 
$\mathbf{F}^{\text{ctx}}_{\text{enc}}$ produced by the 3DST module.
Finally, the 2D-3D interacted and aggregated future voxel features are decoded by either an occupancy head or a flow head to generate the corresponding outputs.

\subsection{2D Spatio-Temporal Module (2DST)}
\label{2DST}
\begin{figure}[t]
    \centering
    \includegraphics[width=1\linewidth]{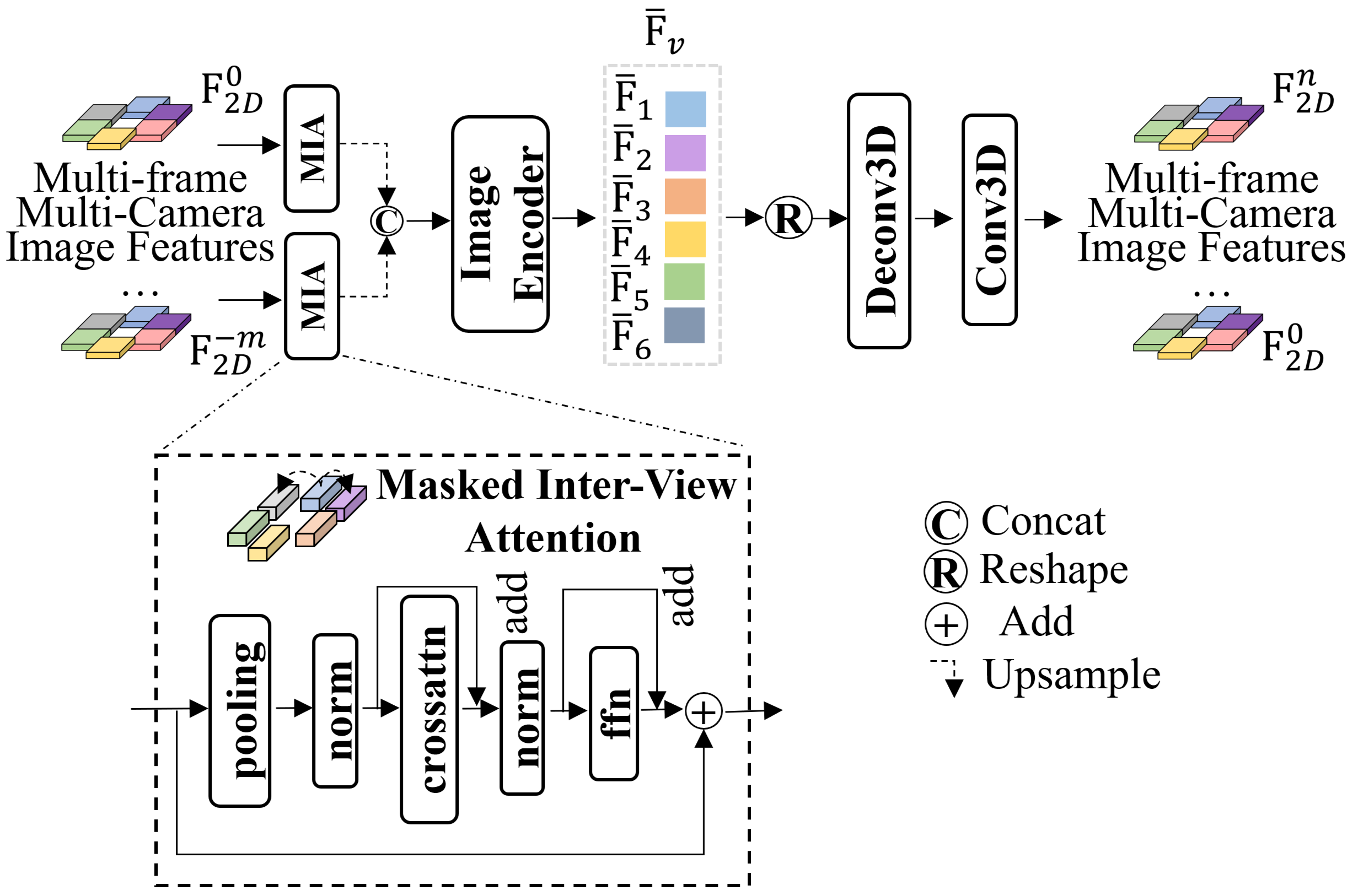}
    \caption {Our 2DST module takes a sequence of historical multi-view image features as input
and predicts future multi-view image features, enabling the forecasting of
future multi-view semantic segmentation through explicit temporal modeling.
}
    \vspace{-0.5cm}
    \label{fig:2DST}
\end{figure}

As illustrated in Fig.~\ref{fig:2DST}, we propose a \textbf{2D Spatio-Temporal (2DST) module} that introduces an auxiliary future segmentation prediction task based on historical multi-view image features $\{\mathbf{F}^{t}_{2D}\}_{t=-m}^{0}$. 
The segmentation supervision provides rich semantic cues and serves as a temporal regularizer, complementing the geometric reasoning in the 3D branch.
Given the boundary-level overlap between adjacent views, we restrict interaction to neighboring cameras and introduce a 
\textit{Masked Inter-view Attention} (\textbf{MIA}) mechanism.
Let $\mathbf{F}_{2D}^{t,v} \in \mathbb{R}^{C \times h \times w}$ denote the feature of view $v$ at time $t$. 
We first compress each feature map along the width dimension to obtain a column-wise representation:
\begin{equation}
\tilde{\mathbf{F}}_{2D}^{t,v}
=
\mathrm{Pool}_w(\mathbf{F}_{2D}^{t,v})
\in
\mathbb{R}^{C \times h \times 1}.
\end{equation}
Based on these pooled features, masked cross-view attention is performed so that each view interacts only with itself and its adjacent views:
\begin{equation}
\hat{\mathbf{F}}_{2D}^{t,v}
=
\mathrm{Attention}\!\left(
\tilde{\mathbf{F}}_{2D}^{t,v},
\{\tilde{\mathbf{F}}_{2D}^{t,u}\}_{u \in \mathcal{N}(v)} 
\right)\in
\mathbb{R}^{C \times h \times 1}.
\end{equation}
where $\mathcal{N}(v)$ denotes the neighboring views of view $v$. 
The attention mask restricts information exchange to adjacent cameras, leveraging their overlapping fields of view to enable more relevant cross-view interactions. The refined column-wise representation $\hat{\mathbf{F}}_{2D}^{t,v}$ is broadcast along the spatial dimension to match the original image representation and injected back via a residual connection. The resulting features are subsequently upsampled and reshaped by merging the temporal and channel dimensions, and fed into a lightweight backbone (e.g., ResNet-18) to produce fused representations $\{\bar{\mathbf{F}}_{2D}^{v}\}_{v=1}^{V}$,
where $\bar{\mathbf{F}}_{2D}^{v} \in \mathbb{R}^{C \times h \times w}$ denotes the fused feature of view $v$.

These fused features are subsequently utilized by the Spatio-Temporal Interaction Module (STIM) (Sec.~\ref{STIM}) for cross-domain feature interaction. In parallel, they are further expanded along the channel dimension and reshaped to recover the future temporal dimension. A 3D transposed convolution network then progressively upsamples the features while preserving spatio-temporal correlations, followed by a 3D convolutional refinement module to enhance temporal consistency. 
Finally, the network produces the predicted future multi-view image features $\{\mathbf{F}_{2D}^{t}\}_{t=0}^{n}$.

By jointly modeling cross-view interactions and temporal dynamics, the 2DST module captures semantic evolution across multi-view observations while maintaining spatial consistency between cameras. 
Compared with independently processing each view, the proposed masked inter-view interaction enables efficient information exchange among adjacent cameras without introducing excessive computational overhead. 
Moreover, the auxiliary future segmentation prediction task provides explicit semantic supervision that complements the geometric reasoning of the 3D voxel branch, leading to more robust spatio-temporal representations for 4D occupancy forecasting.

\subsection{3D Spatio-Temporal Module (3DST)}

\begin{figure}[t]
    \centering
    \includegraphics[width=0.72\linewidth]{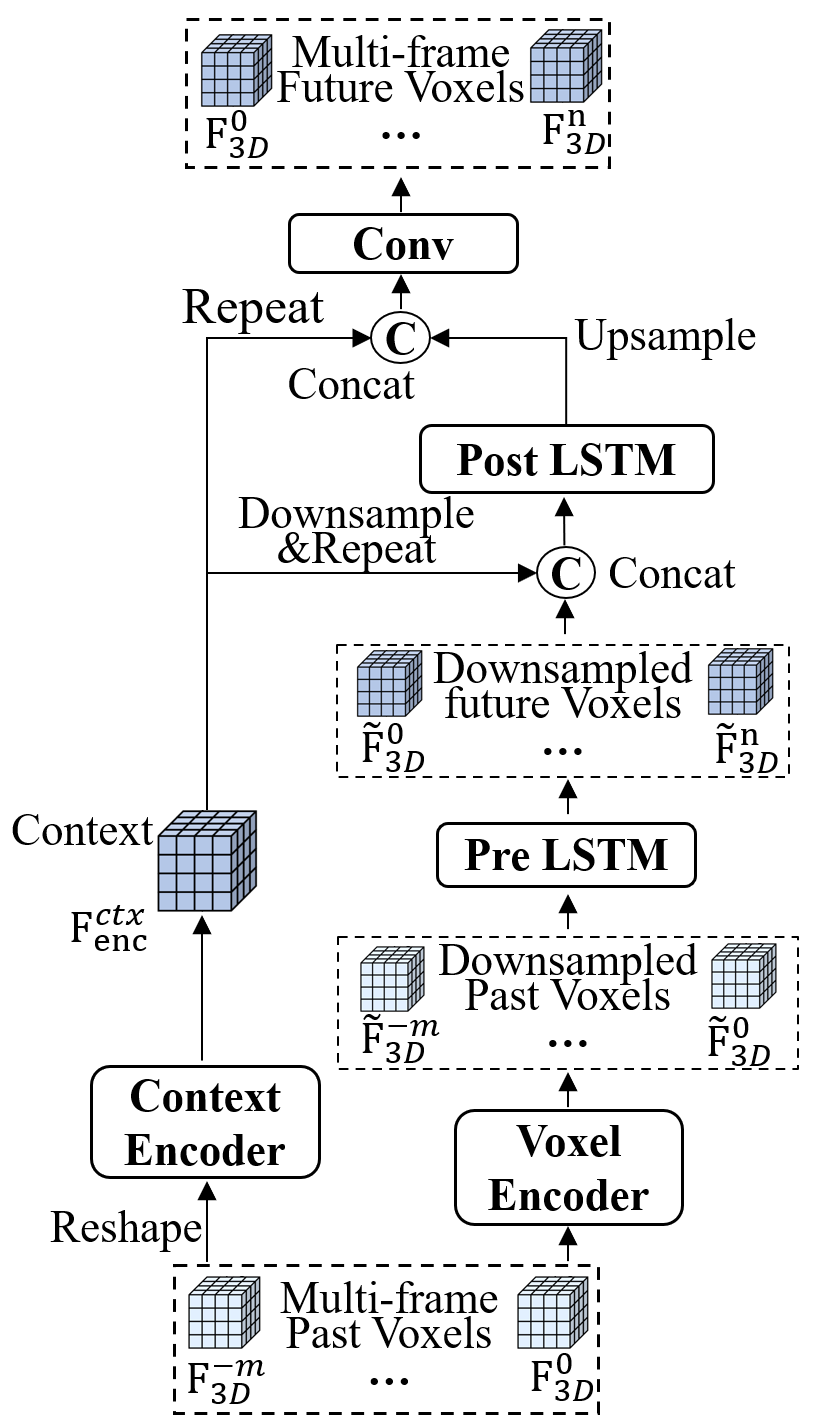} 
    
    \caption{Our 3DST module takes historical voxel features as input, aiming to capture the temporal dynamics. By modeling the evolution of voxel representations, it effectively encodes motion patterns and spatio-temporal context, and further extrapolates them to predict future voxel features.}

    \label{fig:3DST}
\end{figure}

We propose a novel \textbf{3D Spatio-Temporal (3DST) module} that learns volumetric dynamics from historical voxel states to predict future voxel states (see Fig.~\ref{fig:3DST}). This module models motion in a low-resolution space while leveraging high-resolution contextual features to preserve fine-grained spatial details. The input to the 3DST module is a sequence of historical voxel features obtained from the Lifting Module, denoted as $\{\mathbf{F}^{t}_{3D}\}_{t=-m}^{0}$, where $\mathbf{F}^{t}_{3D} \in \mathbb{R}^{C \times H \times W \times D}$.

In the low-resolution branch, we employ a ResNet3D-based voxel encoder $\phi(\cdot)$ to extract multi-scale features and align them to half the spatial resolution, yielding downsampled voxel representations:
\begin{equation}
\{\tilde{\mathbf{F}}^{t}_{3D}\}_{t=-m}^{0} = \phi(\{\mathbf{F}^{t}_{3D}\}_{t=-m}^{0}), 
\end{equation}
where $
\tilde{\mathbf{F}}^{t}_{3D} \in \mathbb{R}^{C\times \frac{H}{2} \times \frac{W}{2} \times \frac{D}{2}}.$ To model the temporal dynamics, the resulting low-resolution voxel sequence is processed by a Pre-LSTM module. It operates autoregressively within a sliding window of size $T$ (where $T=m+1$ is the number of input frames), using the $T$ input frames to predict the next future frame in each step, thereby generating the sequence of future downsampled voxel states:

\begin{equation}
\{\tilde{\mathbf{F}}^{t}_{3D}\}_{t=0}^{n} = \mathrm{LSTM}_{\text{pre}}\left(\{\tilde{\mathbf{F}}^{t}_{3D}\}_{t=-m}^{0}\right).
\end{equation}

In the high-resolution branch, to compensate for the loss of fine-grained spatial details in the low-resolution branch caused by downsampling and temporal propagation, we introduce a context branch that extracts high-resolution spatial context from the original voxel sequences $\{\mathbf{F}^{t}_{3D}\}_{t=-m}^{0}$.
Specifically, the input voxel features $\{\mathbf{F}^{t}_{3D}\}_{t=-m}^{0}$ are first reshaped by stacking the temporal and channel dimensions, and then processed by a context encoder $\psi(\cdot)$ based on ResNet3D~\citep{he2016deep} to produce contextual representations:
\begin{equation}
\mathbf{F}^{\text{ctx}}_{\text{enc}} =
\psi\left(
\mathrm{Reshape}\left(\{\mathbf{F}^{t}_{3D}\}_{t=-m}^{0}\right)
\right)
\in \mathbb{R}^{C \times H \times W \times D}. 
\end{equation}
To incorporate fine-grained information, the encoded contextual feature 
$\mathbf{F}^{\text{ctx}}_{\text{enc}}$ is first fed into the STIM module 
(Sec.~\ref{STIM}) to interact with the fused features $\{\bar{\mathbf{F}}_{2D}^{v}\}_{v=1}^{V}$ from the 2DST branch 
(Sec.~\ref{2DST}), producing a refined contextual representation 
$\hat{\mathbf{F}}^{\text{ctx}}_{\text{enc}}$.
The refined feature is temporally repeated and spatially downsampled to match the resolution of the future voxel features $\{\tilde{\mathbf{F}}^{t}_{3D}\}_{t=0}^{n}$, and is then concatenated with them and fed into a Post-LSTM to jointly enhance temporal coherence and spatial details:

\begin{equation}
\{\hat{\mathbf{F}}^{t}_{3D}\}_{t=0}^{n}
=
\mathrm{LSTM}_{\text{post}}
\left(
\left[
\{\tilde{\mathbf{F}}^{t}_{3D}\}_{t=0}^{n},
\hat{\mathbf{F}}^{\text{ctx}}_{\text{enc}}
\right]
\right),
\end{equation}
where $[\cdot,\cdot]$ denotes channel-wise concatenation and $\hat{F}^{t}_{3D} \in \mathbb{R}^{C\times \frac{H}{2} \times \frac{W}{2} \times \frac{D}{2}}$.
The refined features $\{\hat{\mathbf{F}}^{t}_{3D}\}_{t=0}^{n}$ are finally
upsampled to the original resolution and concatenated with the contextual
features $\mathbf{F}^{\text{ctx}}_{\text{enc}}$ in a frame-wise manner,
followed by a 3D convolutional decoder to produce the future voxel predictions
$\{\mathbf{F}^{t}_{3D}\}_{t=0}^{n}$.

By performing temporal modeling on low-resolution voxel representations and progressively incorporating high-resolution contextual information, the proposed 3DST module benefits from enlarged receptive fields in the low-resolution branch to capture global motion patterns, while preserving fine-grained spatial structures in the high-resolution branch, thereby enabling robust and computationally efficient multi-frame future voxel prediction.

\subsection{Spatio-Temporal Interaction Modeling}
\label{STIM}
\begin{figure}[t]
    \centering
    \includegraphics[width=0.9\linewidth]{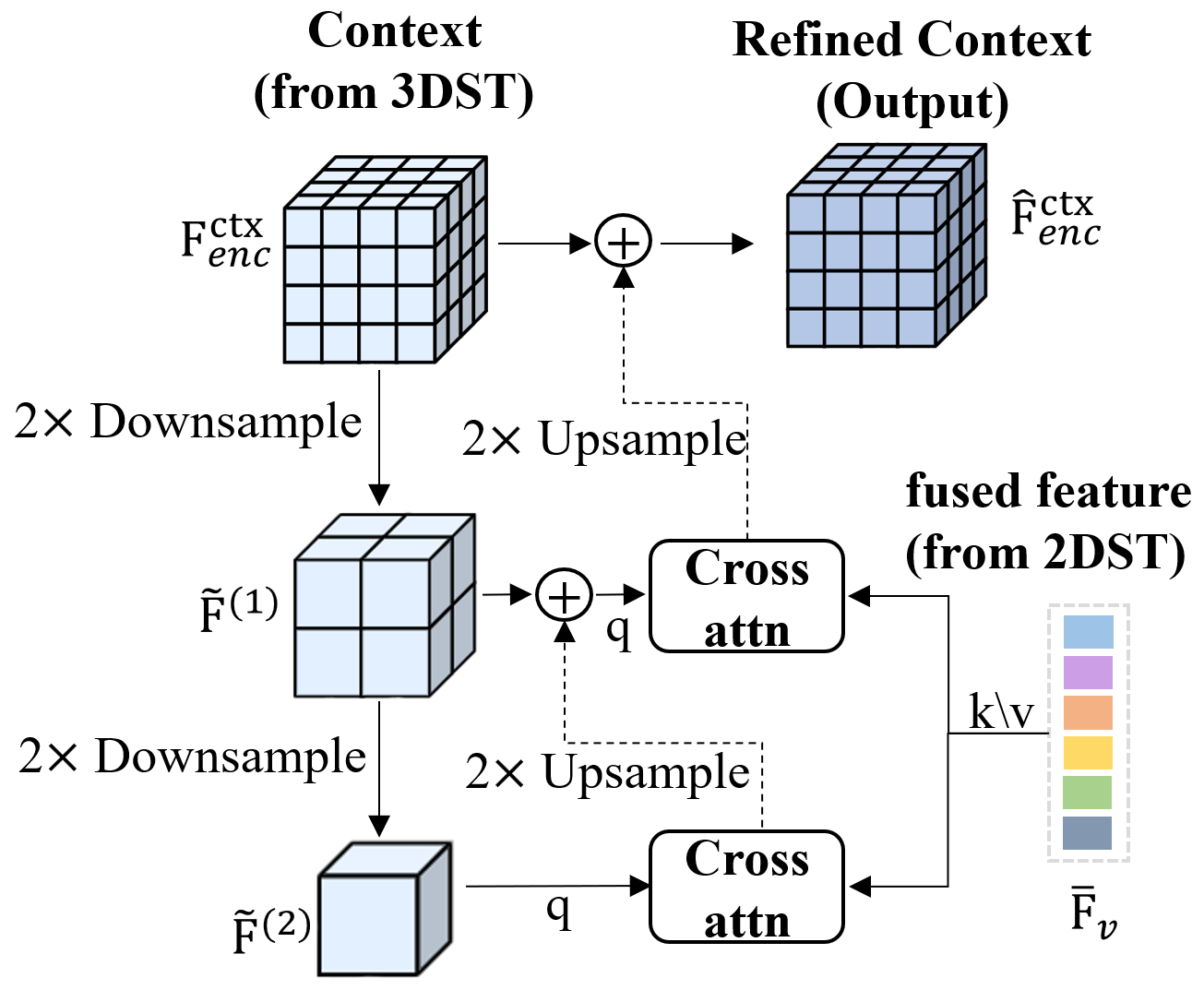}

    \caption {Our STIM module takes the context feature from 3DST and the fused feature from 2DST as inputs, and performs cross-attention interactions between multi-scale context features and the fused feature.
}
    \label{fig:STIM}
\end{figure}

We propose the \textbf{Spatio-Temporal Interaction Modeling (STIM)} module to bridge 2D and 3D feature representations. 
As illustrated in Fig.~\ref{fig:STIM}, STIM takes the contextual feature 
$\mathbf{F}^{\text{ctx}}_{\text{enc}}$ from the 3DST module and the fused multi-view feature 
$\{\bar{\mathbf{F}}_{v}\}_{v=1}^{V}$ from the 2DST module as inputs.
First, the contextual feature 
$\mathbf{F}^{\text{ctx}}_{\text{enc}} \in \mathbb{R}^{C \times H \times W \times D}$
is progressively downsampled twice to construct a two-level feature hierarchy:
\begin{equation}
\begin{aligned}
\tilde{\mathbf{F}}^{(1)} &= \mathcal{D}_1(\mathbf{F}^{\text{ctx}}_{\text{enc}})
\in \mathbb{R}^{C \times \frac{H}{2} \times \frac{W}{2} \times \frac{D}{2}}, \\
\tilde{\mathbf{F}}^{(2)} &= \mathcal{D}_2(\tilde{\mathbf{F}}^{(1)})
\in \mathbb{R}^{C \times \frac{H}{4} \times \frac{W}{4} \times \frac{D}{4}} .
\end{aligned}
\end{equation}
Here, $\mathcal{D}_1(\cdot)$ and $\mathcal{D}_2(\cdot)$ denote 3D downsampling operators that enlarge the receptive field while reducing spatial resolution.
Interaction starts from the lowest-resolution contextual feature.
The feature $\tilde{\mathbf{F}}^{(2)}$ first interacts with the fused multi-view
features through a cross-attention operation, where the contextual feature
serves as the query and the fused image features provide the keys and values:
\begin{equation}
\hat{\mathbf{F}}^{(2)} =
\mathrm{Attn}\!\left(
\tilde{\mathbf{F}}^{(2)},
\mathrm{Flatten}\!\left(\{\bar{\mathbf{F}}_{v}\}_{v=1}^{V}\right)
\right)
\in \mathbb{R}^{C \times \frac{H}{4} \times \frac{W}{4} \times \frac{D}{4}} .
\end{equation}

The resulting feature is then upsampled and merged with the higher-resolution
contextual feature $\tilde{\mathbf{F}}^{(1)}$ through a residual connection:
\begin{equation}
\mathbf{H} =
\tilde{\mathbf{F}}^{(1)} +
\mathcal{UP}\!\left(\hat{\mathbf{F}}^{(2)}\right),
\end{equation}
where $\mathcal{UP}(\cdot)$ denotes the upsampling operation and 
$\mathbf{H} \in \mathbb{R}^{C \times \frac{H}{2} \times \frac{W}{2} \times \frac{D}{2}}$.
The updated feature $\mathbf{H}$ further interacts with the fused multi-view
features via another cross-attention operation. The output is then upsampled
and added back to the original contextual feature to obtain the refined representation:
\begin{equation}
\hat{\mathbf{F}}^{\text{ctx}}_{\text{enc}} =
\mathbf{F}^{\text{ctx}}_{\text{enc}} +
\mathcal{UP}\!\left(
\mathrm{Attn}\!\left(
\mathbf{H},
\mathrm{Flatten}\!\left(\{\bar{\mathbf{F}}_{v}\}_{v=1}^{V}\right)
\right)
\right).
\end{equation}

This bottom-up interaction strategy enables multi-scale cross-modal fusion by progressively integrating complementary information from both modalities. Specifically, low-resolution global context is gradually enriched with detailed appearance cues from the 2D branch, while the high-resolution spatial structure is effectively preserved. Such a design facilitates a coarse-to-fine refinement process, leading to more accurate and structurally consistent representations.

\section{Experiments and Results}
\subsection{Experiment Settings}

\textbf{Datasets.}
Our experiments are conducted on nuScenes~\citep{Caesar_2020_CVPR}, nuScenes-Occupancy~\citep{wang2023openoccupancy}, and Lyft-Level5~\citep{houston2021one} datasets, following the setting of Cam4DOcc. For nuScenes and nuScenes-Occupancy, 700 scenes are selected from the total 850 annotated scenes for training, while the remaining scenes are used for evaluation. Similarly, 130 out of 180 annotated scenes from Lyft-Level5 are used for training, with the rest reserved for testing. This results in 23,930 training sequences and 5,119 evaluation sequences for nuScenes and nuScenes-Occupancy, and 15,720 training sequences and 5,880 evaluation sequences for Lyft-Level5. Each sequence contains 7 frames in total, comprising 3 observation frames (two past frames and the current frame) and 4 future frames for occupancy forecasting. The occupancy prediction region spans $[-51.2\,\mathrm{m}, 51.2\,\mathrm{m}]$ in the horizontal plane and $[-5\,\mathrm{m}, 3\,\mathrm{m}]$ along the vertical axis. A voxel size of $0.2\,\mathrm{m}$ is adopted, leading to a volumetric grid resolution of $(512, 512, 40)$. Due to differences in annotation frequencies across datasets, evaluation results are reported at 2~Hz for nuScenes and nuScenes-Occupancy, and at 5~Hz for Lyft-Level5.

\textbf{Evaluation Protocol and Metrics.}
We follow the Cam4DOcc benchmark to assess camera-only occupancy forecasting under three task settings with increasing annotation granularity. 
(1) \textit{Inflated general movable object (GMO) forecasting:} regions enclosed by annotated 3D bounding boxes in nuScenes and Lyft-Level5 are treated as GMO, while all remaining voxels are assigned to the background category. 
(2) \textit{Fine-grained GMO forecasting:} the category definition remains identical to task (1), but coarse bounding-box annotations are replaced by voxel-level occupancy labels provided by nuScenes-Occupancy. 
(3) \textit{Fine-grained GMO and general static object (GSO) forecasting:} both movable and static object categories are annotated with fine-grained voxel-wise labels. 
Since Lyft-Level5 does not provide fine-grained occupancy annotations, only task (1) is evaluated on this dataset. For all tasks, intersection-over-union (IoU) is used as the evaluation metric, including IoU for the current frame ($\mathrm{IoU}_c$), averaged IoU over future frames ($\mathrm{IoU}_f$), and IoU over the entire prediction horizon ($\mathrm{Io\widetilde{U}}_f$). The evaluation strictly follows the protocol defined in Cam4DOcc.

\textbf{Implementation and Training Details.}
Following OccProphet~\citep{chen2025occprophet}, our method adopts a ResNet-34 backbone~\citep{he2016deep} enhanced with deformable convolutions~\citep{dai2017deformable} as the image encoder, while retaining the same FPN and lifting module. The input image resolution is downsampled to $448 \times 896$. Multi-view 2D semantic segmentation labels are automatically generated using the Segment Anything Model (SAM)~\citep{kirillov2023segment}, thereby eliminating the need for manual annotation; further details can be found in OccNeRF~\citep{zhang2023occnerf}. All ablation studies are conducted on the Lyft-Level5 dataset~\citep{houston2021one}, and all experiments are run using 8 NVIDIA RTX 6000 Ada GPUs. We first trained the model without the STIM module for 24 epochs using the Adam optimizer with a batch size of 8 and a learning rate of $3 \times 10^{-4}$. 
Afterward, the STIM module was introduced, and the entire network was further trained for an additional 6 epochs with a reduced learning rate of $3 \times 10^{-5}$.
 Cross-entropy loss is employed for 2D segmentation and 3D occupancy prediction with weights of 0.05 and 0.5, respectively, while a smooth L1 loss with a weight of 0.5 is applied for flow prediction.

\subsection{Main Results}

\textbf{Evaluation on forecasting inflated GMO.} 
We conduct comprehensive comparisons with OpenOccupancy-C \citep{wang2023openoccupancy}, SPC \citep{wei2023surrounddepth,luo2023pcpnet,zhu2021cylindrical}, PowerBEV-3D \citep{ijcai2023p120}, EfficientOCF \citep{xu2025cvpr}, OccProphet \citep{chen2025occprophet} and the OCFNet baseline \citep{ma2024cam4docc} on nuScenes~\citep{Caesar_2020_CVPR} and Lyft-Level5~\citep{houston2021one} benchmarks. 
As shown in Table~\ref{tab:inf-GMO}, our method achieves the best performance across both datasets and all metrics. 

On \textit{nuScenes}, InterOCF improves the previous best OccProphet from 26.94 to 28.73 in $\mathrm{IoU}_f$ and from 29.15 to 30.05 in $\mathrm{Io\widetilde{U}}_f$. 
Compared with the strong OCFNet baseline (26.82 $\mathrm{IoU}_f$, 27.98 $\mathrm{Io\widetilde{U}}_f$), it brings gains of +1.91 $\mathrm{IoU}_f$ and +2.07 $\mathrm{Io\widetilde{U}}_f$. 
When compared with EfficientOCF (26.05 $\mathrm{IoU}_f$, 28.71 $\mathrm{Io\widetilde{U}}_f$), our method achieves further improvements of +2.68 $\mathrm{IoU}_f$ and +1.34 $\mathrm{Io\widetilde{U}}_f$, validating the effectiveness of our spatio-temporal interaction modeling. 

On \textit{Lyft-Level5,} InterOCF also achieves the best results with 43.72 $\mathrm{IoU}_c$, 39.57 $\mathrm{IoU}_f$, and 40.55 $\mathrm{Io\widetilde{U}}_f$. 
Compared with OCFNet (33.56 $\mathrm{IoU}_f$, 34.60 $\mathrm{Io\widetilde{U}}_f$), our method achieves gains of +5.99 $\mathrm{IoU}_f$ and +5.95 $\mathrm{Io\widetilde{U}}_f$. 
When compared with EfficientOCF (35.07 $\mathrm{IoU}_f$, 39.28 $\mathrm{Io\widetilde{U}}_f$), InterOCF shows improvements of +4.50 $\mathrm{IoU}_f$ and +1.27 $\mathrm{Io\widetilde{U}}_f$, outperforming OccProphet by +1.65, +1.65, and +0.44, respectively.
\textit{These results demonstrate that our method effectively captures the motion patterns of coarse-grained moving objects, showcasing strong forecasting capability and generalization ability.}

\begin{table}[t]
\centering
\caption{Performance on forecasting inflated GMO.}
\vspace{4pt}
\setlength{\tabcolsep}{3pt}
\renewcommand{\arraystretch}{1.05}
\begin{tabular}{l|ccc|ccc}
\toprule
 & \multicolumn{3}{c|}{nuScenes} & \multicolumn{3}{c}{Lyft-Level5} \\
\cline{2-7}
Method & $\text{IoU}_{c}$ & $\text{IoU}_{f}$ (2s) & $\mathrm{Io\widetilde{U}}_f$ 
& $\text{IoU}_{c}$ & $\text{IoU}_{f}$ (0.8s) & $\mathrm{Io\widetilde{U}}_f$ \\
\midrule
OpenOccupancy-C & 12.17 & 11.45 & 11.74 & 14.01 & 13.53 & 13.71 \\
SPC & 1.27 & - & - & 1.42 & - & - \\
PowerBEV-3D & 23.08 & 21.25 & 21.86 & 26.19 & 24.47 & 25.06 \\
BEVDet4D & 31.60 & 24.87 & 26.87 & - & - & - \\
OCFNet & 31.30 & 26.82 & 27.98 & 36.41 & 33.56 & 34.60 \\
EfficientOCF & 34.13 & 26.05 & 28.71 & 41.24 & 35.07 & 39.28 \\
OccProphet & 34.36 & 26.94 & 29.15
& 43.38 & 37.92 & 40.11 \\
\hline
\rowcolor[HTML]{F9F2FE}
InterOCF (ours) & \textbf{34.72} & \textbf{28.73} & \textbf{30.05} & \textbf{43.72} & \textbf{39.57} & \textbf{40.55} \\
\bottomrule
\end{tabular}
\label{tab:inf-GMO}
\end{table}

\textbf{Evaluation on forecasting fine-grained GMO.} We further compare our method with the above approaches on the nuScenes-Occupancy~\citep{wang2023openoccupancy} benchmark. 
As shown in Table~\ref{tab:fine-GMO}, InterOCF achieves the best performance across all evaluation metrics. 
Specifically, our method obtains 15.71 $\mathrm{IoU}_c$, 11.63 $\mathrm{IoU}_f$, and 12.12 $\mathrm{Io\widetilde{U}}_f$, outperforming all competing methods. In particular, it improves upon the previous best OccProphet by +0.33 $\mathrm{IoU}_c$, +0.94 $\mathrm{IoU}_f$, and +0.14 $\mathrm{Io\widetilde{U}}_f$. 
When comparing with EfficientOCF (12.31 $\mathrm{IoU}_c$, 9.67 $\mathrm{IoU}_f$, 10.66 $\mathrm{Io\widetilde{U}}_f$), InterOCF shows significant improvements of +3.40 $\mathrm{IoU}_c$, +1.96 $\mathrm{IoU}_f$, and +1.46 $\mathrm{Io\widetilde{U}}_f$. 
Compared to OCFNet (11.45 $\mathrm{IoU}_c$, 9.68 $\mathrm{IoU}_f$, 10.10 $\mathrm{Io\widetilde{U}}_f$), our method demonstrates even larger gains of +4.26 $\mathrm{IoU}_c$, +1.95 $\mathrm{IoU}_f$, and +2.02 $\mathrm{Io\widetilde{U}}_f$. 
\textit{These results demonstrate the effectiveness of our method in capturing fine-grained dynamic motion and producing more accurate future occupancy predictions.}

\begin{table}[t]
\centering

\captionof{table}{Performance on forecasting fine-grained GMO.}
\vspace{4pt}

\renewcommand\arraystretch{1.4}
\setlength{\tabcolsep}{6pt}

\begin{tabular}{l|ccc}
\toprule
 & \multicolumn{3}{c}{nuScenes-Occupancy} \\
\cline{2-4}
\multirow{-2}{*}{Method} & $\text{IoU}_{c}$ & $\text{IoU}_{f}$ (2s) & $\mathrm{Io\widetilde{U}}_f$ \\
\midrule
OpenOccupancy-C~\citep{wang2023openoccupancy} & 10.82 & 8.02 & 8.53\\
SPC~\citep{wei2023surrounddepth,luo2023pcpnet,zhu2021cylindrical} & 5.85 & 1.08 & 1.12\\
PowerBEV-3D~\citep{ijcai2023p120} & 5.91 & 5.25 & 5.49\\
OCFNet (Cam4DOcc)~\citep{ma2024cam4docc} & 11.45 & 9.68 & 10.10\\
EfficientOCF~\citep{xu2025cvpr} &12.31 & 9.67 & 10.66\\
OccProphet~\citep{chen2025occprophet} & 15.38 & 10.69 & 11.98\\
\hline
\rowcolor[HTML]{F9F2FE}
InterOCF (ours) & \textbf{15.71} & \textbf{11.63} & \textbf{12.12}\\
\bottomrule
\end{tabular}

\label{tab:fine-GMO}

\end{table}

\textbf{Evaluation on forecasting fine-grained GMO and fine-grained GSO.}
Our proposed framework establishes strong performance on the challenging task of fine-grained future scene forecasting, significantly outperforming existing methods in predicting both General Moving Objects (GMO) and General Static Occupancy (GSO). The quantitative results, summarized in Table~\ref{tab:fine-GMO_fine-GSO}, demonstrate our method's comprehensive superiority. Our method achieves the highest scores across all critical evaluation metrics: it attains a mean $\text{IoU}_c$ of 19.57 (with a decomposition of 14.24 for GMO and 24.90 for GSO), an $\text{IoU}_f(2s)$ of 17.97 (10.26 for GMO and 25.67 for GSO), and a $\mathrm{Io\widetilde{U}}_f$ of 10.85 specifically for GMO forecasting. 

For fine-grained GMO forecasting, our method outperforms EfficientOCF by +0.72 $\mathrm{IoU}_c$, +1.13 $\mathrm{IoU}_f(2s)$, and +1.03 $\mathrm{Io\widetilde{U}}_f$, and exceeds OCFNet by +3.22 $\mathrm{IoU}_c$, +1.06 $\mathrm{IoU}_f(2s)$, and +1.19 $\mathrm{Io\widetilde{U}}_f$.
For fine-grained static occupancy prediction, our method surpasses EfficientOCF by +0.79 $\mathrm{IoU}_c$ and +6.46 $\mathrm{IoU}_f(2s)$, and outperforms OCFNet by +7.11 $\mathrm{IoU}_c$ and +7.84 $\mathrm{IoU}_f(2s)$.
Notably, the performance gain in $\mathrm{Io\widetilde{U}}_f$ for GMO---a key metric for evaluating the forecasting of dynamic agents---represents a substantial improvement of 5.1\% over the previous leading method, OccProphet \citep{chen2025occprophet}, a marked 12.4\% increase over OCFNet \citep{ma2024cam4docc}, and 10.5\% improvement over EfficientOCF.

\textit{These results demonstrate our method’s strong capability in capturing fine-grained static scene details and dynamic object behaviors, effectively modeling complex spatio-temporal dependencies in autonomous driving scenarios, and leading to more accurate and reliable predictions.}

\begin{table}[t]
\centering

\captionof{table}{Performance on forecasting fine-grained GMO and fine-grained GSO.}
\vspace{4pt}

\renewcommand\arraystretch{1.4}
\setlength{\tabcolsep}{2pt}

\begin{tabular}{l|ccc|ccc|c}
\toprule
 & \multicolumn{3}{c|}{$\text{IoU}_{c}$} & \multicolumn{3}{c|}{$\text{IoU}_{f}$ (2s)} & $\mathrm{Io\widetilde{U}}_f$\\
\cline{2-8}
\multirow{-2}{*}{Method} & GMO & GSO & mean & GMO & GSO & mean & GMO\\
\midrule
OpenOccupancy-C~\citep{wang2023openoccupancy} & 9.62 & 17.21 & 13.42 & 7.41 & 17.30 & 12.36 & 7.86\\
SPC~\citep{wei2023surrounddepth,luo2023pcpnet,zhu2021cylindrical} & 5.85 & 3.29 & 4.57 & 1.08 & 1.40 & 1.24 & 1.12\\
PowerBEV-3D~\citep{ijcai2023p120} & 5.91 & - & - & 5.25 & - & - & 5.49 \\
OCFNet (Cam4DOcc)~\citep{ma2024cam4docc} & 11.02 & 17.79 & 14.41 & 9.20 & 17.83 & 13.52 & 9.66 \\
EfficientOCF~\citep{xu2025cvpr} &13.52 & 24.11 & 18.82 & 9.13 & 19.21 & 14.17 & 9.82\\
OccProphet~\citep{chen2025occprophet} & 13.71 & 24.42 & 19.06 & 9.34 & 24.56 & 16.95 & 10.33 \\
\hline
\rowcolor[HTML]{F9F2FE}
InterOCF (ours) & \textbf{14.24} & \textbf{24.90} & \textbf{19.57} & \textbf{10.26} & \textbf{25.67} & \textbf{17.97} & \textbf{10.85} \\
\bottomrule
\end{tabular}

\label{tab:fine-GMO_fine-GSO}

\end{table}

\subsection{Ablation Study}

\textbf{Module Ablations.}
We conduct ablation studies to evaluate the contributions of key components in our method framework, with results summarized in Table \ref{tab:ablation_study}. 
Starting from the baseline model, we use only a single linear layer to predict future states (Row~1, 37.65 $\text{IoU}_{f}$), 
introducing the full 3DST module improves the performance to 38.04, 
yielding a gain of +0.39. 
When removing the PostLSTM from 3DST, the performance drops to 37.83, 
indicating that temporal refinement plays a non-negligible role in modeling motion consistency.
Building upon 3DST, incorporating the full 2DST module further boosts the performance to 39.09, 
achieving a significant gain of +1.05 over the 3DST-only configuration. 
However, removing the masked inter-view attention (MIA) results in a decrease to 38.44 (-0.65), 
demonstrating the effectiveness of inter-view interaction for enhancing spatial alignment and feature aggregation.
Finally, integrating the proposed STIM module brings additional improvement, 
achieving the best performance of 39.57 $\text{IoU}_{f}$ (+0.48). 
This consistent gain verifies that STIM effectively facilitates cross-domain spatio-temporal feature interaction, 
leading to more accurate future occupancy prediction.

\begin{table}[t]
\centering
\caption{Ablation study on method modules.}
\vspace{4pt}

\small
\renewcommand\arraystretch{1.15}
\setlength{\tabcolsep}{3.5pt}

\begin{tabular}{lccccc c}
\toprule
 & \multicolumn{2}{c}{3DST} & \multicolumn{2}{c}{2DST} & STIM & $\mathrm{IoU}_f$ $\uparrow$ \\
\cmidrule(lr){2-3} \cmidrule(lr){4-5}
Config. & Full & PostLSTM & Full & MIA &  & (0.8s) \\
\midrule

\rowcolor[HTML]{F5F5F5}
Baseline & -- & -- & -- & -- & -- & 37.65 \\

\rowcolor[HTML]{E8F5E9}
+3DST & \checkmark & \checkmark & -- & -- & -- & 38.04 (+0.39) \\

\rowcolor[HTML]{FFF3E0}
w/o Post & \checkmark & $\times$ & -- & -- & -- & 37.83 (-0.21) \\

\rowcolor[HTML]{E8F5E9}
+2DST & \checkmark & \checkmark & \checkmark & \checkmark & -- & 39.09 (+1.05) \\

\rowcolor[HTML]{FFF3E0}
w/o MIA & \checkmark & \checkmark & \checkmark & $\times$ & -- & 38.44 (-0.65) \\

\rowcolor[HTML]{F3E5F5}
+STIM & \checkmark & \checkmark & \checkmark & \checkmark & \checkmark & \textbf{39.57 (+0.48)} \\

\bottomrule
\end{tabular}

\label{tab:ablation_study}
\end{table}

\textbf{Ablation on Frame Count for 2D Prediction.}
\label{2dlength}
We conduct an ablation study on the temporal horizon length of the 2D semantic-temporal (2DST) module's multi-view segmentation predictions. The number of predicted frames in the 2D branch influences how much temporal context is available for 3D temporal prior aggregation. We evaluate our full method on the Lyft-Level5 dataset~\citep{houston2021one} using the forward intersection-over-union $\text{IoU}_{f}$ as the evaluation metric. 

As shown in Table~\ref{tab:frame_count}, performance systematically improves as the 2D prediction length increases. With a single predicted frame, the model achieves 38.92 $\text{IoU}_{f}$. Extending to 3 frames yields 39.23 $\text{IoU}_{f}$, representing a 0.31-point improvement. The best performance of 39.57 $\text{IoU}_{f}$ is attained when the 2D prediction horizon matches the output length of the 3D semantic-temporal module (5 frames). This trend reveals that richer 2D temporal priors directly benefit 4D occupancy forecasting. Shorter horizons provide limited motion context, constraining the 3D module's ability to reason about scene dynamics. The performance plateau with 5 frames suggests that matching the temporal extents of 2D and 3D branches enables optimal cross-modal feature alignment, allowing comprehensive spatiotemporal information flow from 2D predictions to guide 4D occupancy evolution. This configuration ensures that sufficient short-term motion cues are captured in 2D space before being propagated to 3D, validating our design choice of synchronized temporal dimensions across modules.

\textbf{Ablation on temporal model.}
To select the most suitable sequence modeling architecture for our task, we conduct comprehensive experiments comparing several classical and well-established sequence modeling architectures, including ConvLSTM \citep{shi2015convolutional} (here uniformly referred to as LSTM), Transformer~\citep{vaswani2017attention}, and GRU~\citep{cho2014learning} variants. These architectures represent distinct paradigms for sequential data modeling, enabling a rigorous assessment of their suitability for future occupancy forecasting.

Based on the comparative results in Table~\ref{tab:temporal_model}, LSTM achieves the highest $\text{IoU}_{f}$ of 39.57, surpassing Transformer (38.77) and GRU (37.51) by +0.80 and +2.06, respectively. While the Transformer is theoretically capable of modeling long-range dependencies through global attention, its performance is slightly inferior in this specific setting. We hypothesize that the observed performance gap may be attributed to the relatively short temporal horizon of the forecasting task and the limited training scale, which can hinder attention-based models from learning robust temporal representations. In contrast, LSTM introduces a recurrent inductive bias that is well-suited for modeling short- to mid-term temporal dynamics in driving scenes. Its gated memory mechanism facilitates selective propagation of relevant spatio-temporal context, which is beneficial for accurate future occupancy prediction. 
Based on these observations, we adopt LSTM as the default temporal modeling module in our framework.

\begin{table}[t]
\centering
\setlength{\tabcolsep}{6pt}
\renewcommand{\arraystretch}{1.1}

\begin{minipage}{0.48\linewidth}
\centering
\caption{Ablation on frame count.}
\vspace{4pt}
\begin{tabular}{cc}
\toprule
Frame & $\mathrm{IoU}_f$ \\
\midrule
1 & 38.92 \\
3 & 39.23 \\
\rowcolor[HTML]{F9F2FE}
5 & \textbf{39.57} \\
\bottomrule
\end{tabular}
\label{tab:frame_count}
\end{minipage}
\hfill
\begin{minipage}{0.48\linewidth}
\centering
\caption{Ablation on temporal model.}
\vspace{4pt}
\begin{tabular}{cc}
\toprule
Method & $\mathrm{IoU}_f$ \\
\midrule
GRU & 37.51 \\
Transformer & 38.77  \\
\rowcolor[HTML]{F9F2FE}
LSTM & \textbf{39.57 } \\
\bottomrule
\end{tabular}
\label{tab:temporal_model}
\end{minipage}

\end{table}

\begin{figure*}[t]
    \centering
    \includegraphics[width=1\linewidth]{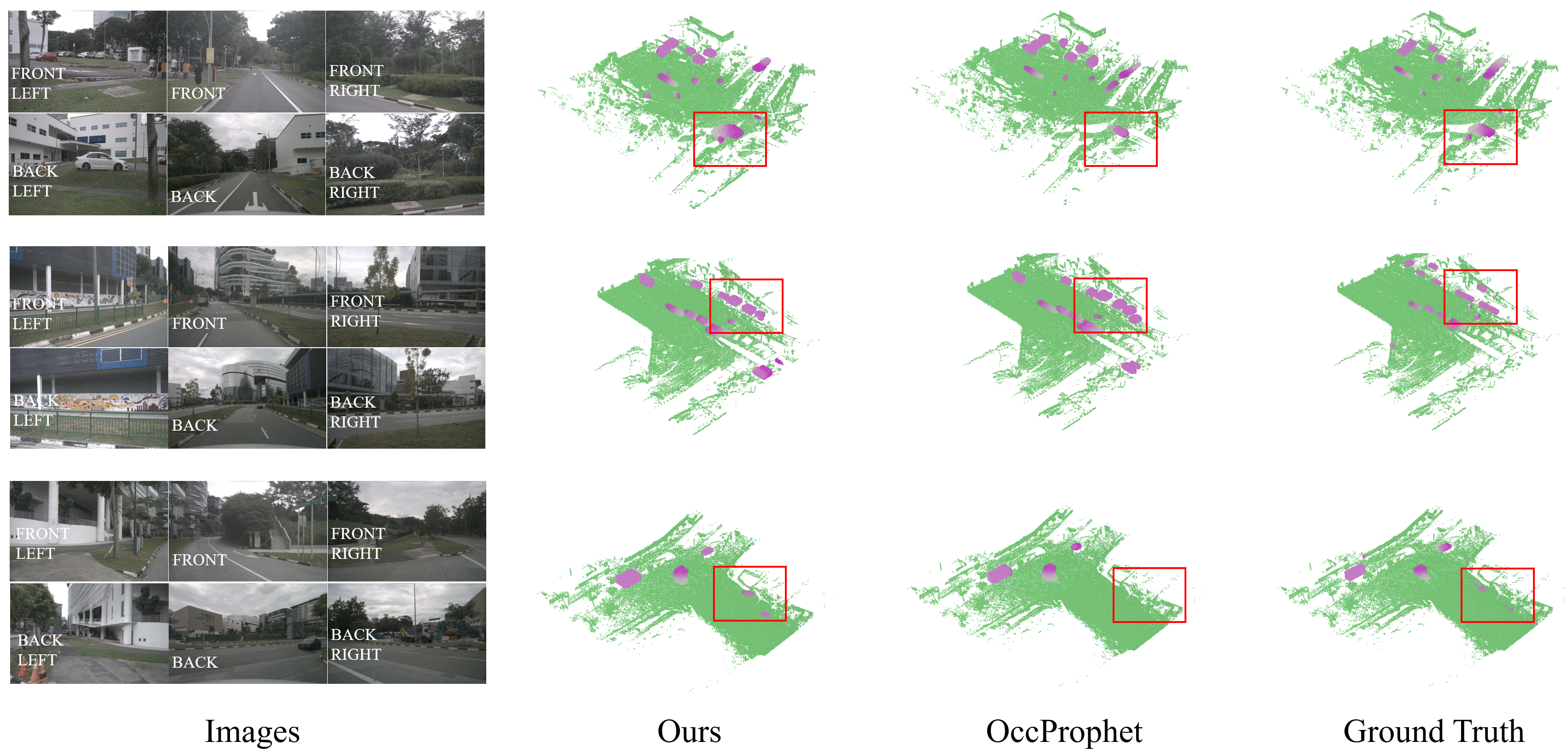}

    \caption {Qualitative results of OccProphet and our InterOCF in the future 2 seconds. The moving objects gradually transitions to white in color along the direction of motion. Red dashed rectangles represent that the results of
our InterOCF are more consistent with the ground truth than those of OccProphet.
}
    \label{fig:VISUAL1}
\end{figure*}

\begin{figure*}[t]
    \centering
    \includegraphics[width=1\linewidth]{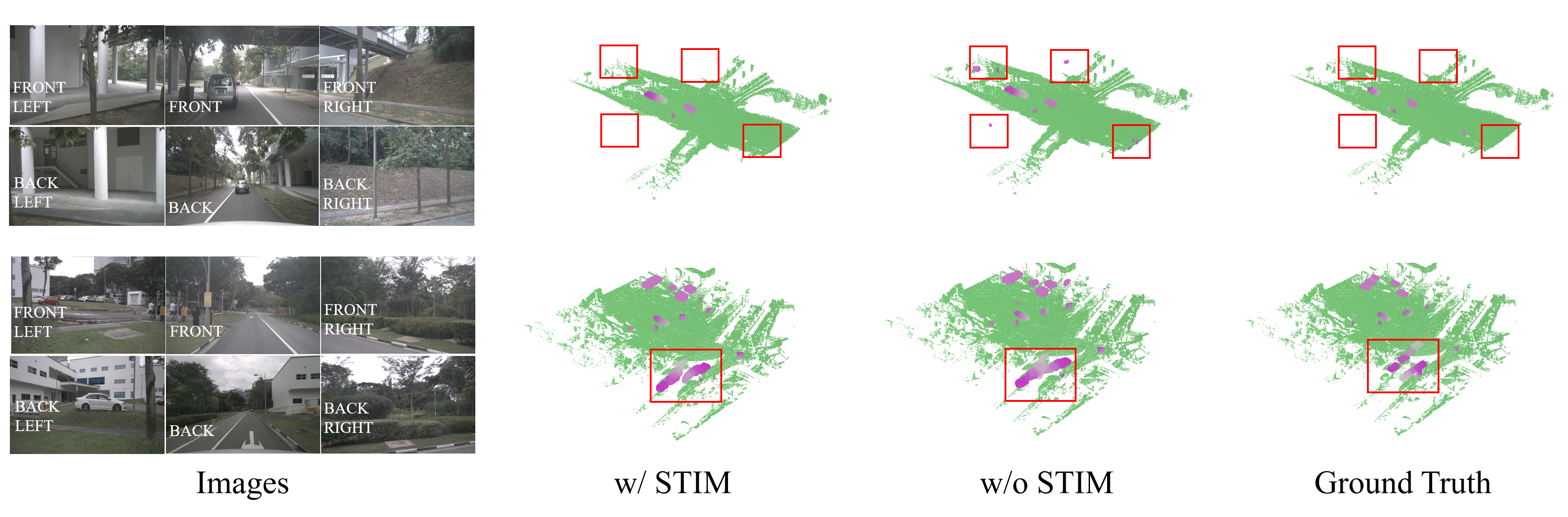}

    \caption {Qualitative results of using the STIM or not. The moving objects gradually transitions to white in color along the direction of motion. Red dashed rectangles represent that the results with the STIM are more
consistent with the ground truth than those without the STIM.
}
    \label{fig:VISUAL2}
\end{figure*}

\subsection{ Visualization}

\textbf{Qualitative Comparison with Baselines.} 
Fig.~\ref{fig:VISUAL1} presents qualitative comparisons between our InterOCF and OccProphet~\citep{chen2025occprophet}. 
Given multi-view camera images as input, we visualize the predicted future occupancy and compare it with the ground truth. Moving objects gradually transition to white along their motion trajectories. As highlighted in the red boxes, our method produces predictions that are more consistent with the ground truth, particularly in regions containing dynamic objects.

In the first scene, our method accurately captures both moving objects within the red bounding boxes and correctly infers their motion directions. Notably, the larger object is predicted to turn. In contrast, OccProphet misses the smaller object and incorrectly predicts the larger one as moving straight ahead, failing to capture its turning behavior. 
In the second scene, the ground truth shows that stationary vehicles along the roadside are parked longitudinally. OccProphet incorrectly predicts them as laterally parked, whereas our predictions better reflect realistic vehicle layouts.
In the third scene, which involves long-range and sparse objects, OccProphet tends to miss or distort distant vehicles. In comparison, our method better preserves geometric consistency and produces predictions that are more aligned with the ground truth.
These results demonstrate the effectiveness of our spatio-temporal modeling strategy, enabling more reliable motion reasoning and more accurate future occupancy prediction.

\textbf{Qualitative Comparison of Using STIM vs. Not Using STIM.}
To further validate the effectiveness of our interaction modeling, we present qualitative ablation results of the STIM module in Fig.~\ref{fig:VISUAL2}. 
In the first scene, the model equipped with STIM produces predictions that are closer to the ground truth, as highlighted in the red boxes. In contrast, the model without STIM exhibits noticeable noise and artifacts, indicating that our interaction modeling improves robustness and noise resistance.
In the second scene, the predictions with STIM are more physically plausible compared to those without STIM, avoiding unrealistic trajectory collisions. This demonstrates that, by leveraging cross-modal interactions, our approach yields more reliable and coherent predictions.

\subsection{Comparison of Model complexity}
We further compare the computational complexity of our proposed InterOCF with several representative methods. As shown in Table~\ref{tab:comp_flops}, InterOCF achieves a favorable trade-off between forecasting accuracy and computational efficiency.
Compared with OccProphet, InterOCF improves $\mathrm{IoU}_f$ from 37.92 to 39.57 (+1.65) while reducing GPU memory consumption from 24\,G to 22\,G. Meanwhile, the inference speed remains comparable (4.5 vs. 4.3 FPS), indicating that the moderate increase in parameters and FLOPs is effectively translated into tangible performance gains.
Compared with earlier heavy 3D methods such as Cam4DOcc (6434\,G FLOPs, 57\,G memory), InterOCF significantly reduces redundant voxel-level computation. This efficiency mainly arises from our hierarchical design, where global motion patterns are modeled in a compact low-resolution voxel space, while fine-grained structural details are preserved through high-resolution feature refinement. As a result, InterOCF achieves higher forecasting accuracy with substantially lower memory consumption and competitive runtime efficiency, making it more practical for real-world deployment.

\begin{table}[t]
\centering
\vspace{4pt}
\caption{Comparison of model complexity. N$_{\text{p}}$: Total number of parameters. Mem.: Occupied GPU memory during training.}
\renewcommand\tabcolsep{1pt}

    \begin{tabular}{c|ccccc}
    \toprule
    Method & N$_{\text{p}}$ (M) $\downarrow$ & Mem. (G) $\downarrow$ & FLOPs (G) $\downarrow$ & FPS $\uparrow$ & $\text{IoU}_{f}$ $\uparrow$ \\
    \midrule
    Cam4DOcc & 370 & 57 & 6434 & 1.7 & 33.56 \\
     EfficientOCF & 393 & $>$46 & - & 3.7 & 35.07 \\
     OccProphet & \textbf{82} & \underline{24} & \textbf{1985} & \textbf{4.5} & \underline{37.92}\\

     \rowcolor[HTML]{F9F2FE}
    InterOCF (Ours) & \underline{112} & \textbf{22} & \underline{2436} & \underline{4.3} & \textbf{39.57}\\
    \bottomrule
    \end{tabular}

\label{tab:comp_flops}
\vspace{-7pt}
\end{table}

\section{Discussion and Conclusion}
In this work, we study the problem of camera-only 4D occupancy forecasting, which aims to predict future 3D semantic scenes from historical multi-view images for autonomous driving. To address the insufficient spatio-temporal modeling in existing methods, we propose a novel framework, \textbf{InterOCF}, that jointly models temporal dynamics in both 3D voxel representations and multi-view segmentation sequences while enabling explicit interaction between the 2D and 3D branches. InterOCF consists of three key components: the \textbf{3D Spatio-Temporal (3DST)} module for modeling volumetric dynamics from historical voxel states, the \textbf{2D Spatio-Temporal (2DST)} module that introduces an auxiliary multi-view temporal segmentation forecasting task to enhance semantic temporal modeling, and the \textbf{Spatio-Temporal Interaction Modeling (STIM)} module that facilitates feature interaction between the 2D and 3D representations. Extensive experiments on the nuScenes, Lyft-Level5, and nuScenes-Occupancy datasets demonstrate that InterOCF consistently outperforms existing baselines, validating the effectiveness of jointly modeling spatio-temporal dynamics in both 2D and 3D domains, as well as their interactions, for accurate 4D occupancy forecasting.


\section*{Acknowledgements}
This work was supported in parts by NSFC (62202312), Guangdong Basic and Applied Basic Research Foundation (2023B1515120026), Shenzhen Science and Technology Program (KQTD 20210811090044003, RCJC20200714114435012), and Scientific Foundation for Youth Scholars and Scientific Development Funds of Shenzhen University.

\bibliographystyle{IEEEtran}
\bibliography{ref}

\end{document}